\documentclass[10pt,twocolumn,letterpaper]{article}

\usepackage{cvpr2019}
\usepackage{times}
\usepackage{epsfig}
\usepackage{graphicx}
\graphicspath{{./Figures/}}
\usepackage{amsmath}
\usepackage{amssymb}

\usepackage{enumitem}
\usepackage[utf8]{inputenc}
\usepackage{amsmath}
\usepackage{amsfonts}
\usepackage{amssymb}
\usepackage{multirow}
\usepackage{color}
\usepackage{bm}
\usepackage{placeins}
\usepackage{pifont}

\newcommand{\cmark}{\ding{51}}
\newcommand{\xmark}{\ding{55}}

\usepackage[pagebackref=true,breaklinks=true,letterpaper=true,colorlinks,bookmarks=false]{hyperref}

\cvprfinalcopy 

\def\cvprPaperID{1870} 

\ifcvprfinal\fi
\begin{document}

\title{Fast Human Pose Estimation}

\author{Feng Zhang$^1$ 
	\quad \quad  \quad \quad  \quad Xiatian Zhu$^2$ 
	\quad \quad \quad \quad \quad  Mao Ye$^{1}$ \\
	$^1${\tt\small \{zhangfengwcy, cvlab.uestc\}@gmail.com},
	School of Computer Science and Engineering, \\
	University of Electronic Science and Technology of China \\
	$^2${\tt\small eddy@visionsemantics.com}, Vision Semantics Limited  \\
}



\maketitle
\thispagestyle{empty}

\begin{abstract}
   Existing human pose estimation approaches 
   often only consider how to improve the model generalisation performance,
   but putting aside the significant efficiency problem.
   This leads to the development of 
   heavy models with poor scalability 
   and cost-effectiveness in practical use.
   In this work, we investigate 
   the under-studied but practically critical pose model efficiency problem.
   To this end, 
   we present a new {\em Fast Pose Distillation} (FPD) model learning 
   strategy.
   Specifically, the FPD trains a lightweight pose neural network architecture
   capable of executing rapidly with low computational cost.
   It is achieved 
   by effectively transferring the pose structure knowledge of a strong teacher
   network.
   Extensive evaluations demonstrate the advantages of 
   our FPD method over a broad range of state-of-the-art
   pose estimation approaches in terms of model cost-effectiveness
   on two standard benchmark datasets,
   MPII Human Pose and Leeds Sports Pose.
\end{abstract}

\section{Introduction}

Human pose estimation has gained remarkable progress 
from the rapid development of 
various deep CNN models
\cite{toshev2014deeppose,carreira2015human,chen2017adversarial}. 
This is because deep neural networks 
are strong at approximating 
complex and non-linear mapping functions from
arbitrary person images to the joint locations 
even at the presence of unconstrained
human body appearance, viewing conditions and background noises.

Nevertheless, the model performance advantages come with the cost of 
training and deploying
resource-intensive networks with large depth and width.
This causes inefficient model inference, 
requiring per-image computing cost at tens of FLoating point OPerations (FLOPs)
therefore poor scalability particularly on resource-limited devices such as 
smart phones and robots.
There is a recent attempt 
that binarises the network parameters 
for model execution speedup \cite{bulat2017binarized},
which however suffers significantly weak model generalisation capacity.

In this study, we consider the problem of
improving the pose estimation efficiency 
{\em without} model performance degradation
but preserving comparable accuracy results.
We observe that the basic CNN building blocks
for state-of-the-art human pose networks such as 
Hourglass \cite{newell2016stacked} are not cost-effective 
in establishing small networks due to a high number of channels per layer
and being more difficult to train.
To overcome these barriers, we design a lightweight variant of Hourglass
network and propose a more effective training method of small pose networks
in a knowledge distillation fashion \cite{hinton2015distilling}.
We call the proposed method \textit{Fast Pose Distillation} (FPD).
Compared with the top-performing alternative pose approaches \cite{yang2017pyramid,chen2017adversarial},
the proposed FPD approach enables much faster and more cost-effective model inference with extremely smaller model size
while simultaneously reaching the same level of human pose prediction performance.

We summarise our {\em contributions} in follows:
\begin{enumerate}[label=(\roman*)]
	\item We investigate the under-studied human pose model efficiency
	problem, opposite to
	the existing attempts mostly focusing on improving
	the accuracy performance {\em alone} at high costs of 
	model inference at deployment.
	This is a critical problem to be addressed for scaling up the existing deep pose estimation methods to real applications.
	
	\item We propose a {\em Fast Pose Distillation} (FPD) model
	training method enabling to more effectively train extremely small 
	human pose CNN networks.
	This is based on an idea of knowledge distillation
	that have been successfully exploited in inducing 
	object image categorisation deep models.
	In particular, we derive a pose knowledge distillation learning objective
	to transfer the latent knowledge from a pre-trained larger teacher model
	to a tiny target pose model (to be deployed in test time).
	This aims to pursue the best model performance 
	given very limited computational budgets 
	using only a small fraction (less than 20\%) of cost required by similarly strong 
	alternatives.
	
	\item We design a lightweight Hourglass network capable of 
	constructing more cost-effective pose estimation CNN models while
	retaining sufficient learning capacity for allowing satisfactory
	accuracy rates.
	This is achieved by extensively examining the redundancy degree of existing
	state-of-the-art pose CNN architecture designs.
\end{enumerate}

In the evaluations, we have conducted extensive empirical comparisons to validate the
efficacy and superiority of the proposed FPD method over a wide
variety of state-of-the-art human pose estimation approaches in the balance
of model inference efficiency and prediction performance
on two commonly adopted
benchmark datasets, MPII Human Pose \cite{andriluka14cvpr} and
Leeds Sports Pose \cite{johnson2010clustered}.

\begin{figure*}[h]
	\centering
	\includegraphics[width=1\linewidth]{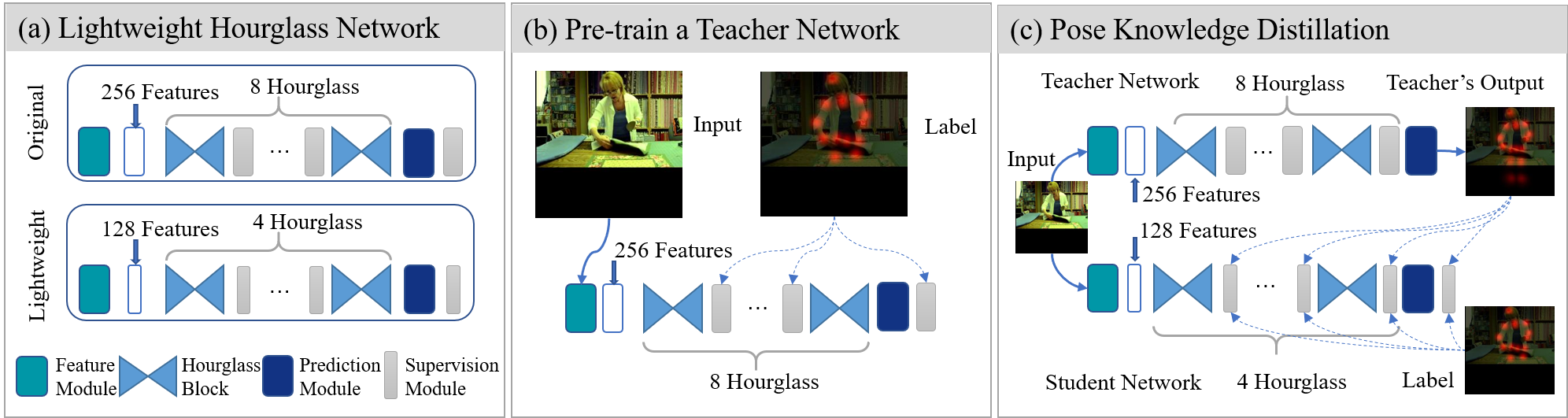}
	\vskip 0.1cm
	\caption{
		An overview of the proposed Fast Pose Distillation model learning strategy.
		To establish a highly cost-effective human pose estimation model, 
		We need to build a compact backbone such as {\bf (a)} a lightweight Hourglass network.
		To more effectively train a small target network, we adopt the principle of
		knowledge distillation in the pose estimation context. 
		This requires to {\bf (b)} pre-train a strong teacher pose model, such as the state-of-the-art Hourglass network or other existing alternatives.
		The teacher model is used to provide extra supervision guidance in the {\bf (c)} pose knowledge distillation procedure via the proposed mimicry loss function. 
		At test time, the small target pose model enables a fast and cost-effective deployment. The computationally expensive teacher model is {\em abandoned} finally,
		since its discriminative knowledge transferred already into the target model therefore
		used in deployment (rather than wasted).
	}
	\label{fig:pipeline}
\end{figure*}

\section{Related Work}
{\bf Human Pose Estimation } 
The past five years have witnessed a huge progress of 
human pose estimation in the deep learning regime
\cite{toshev2014deeppose,tompson2015efficient,bulat2016human,wei2016convolutional,newell2016stacked,chu2017multi,yang2017pyramid,Nie_2018_CVPR,Peng_2018_CVPR}.
Despite the clear performance increases, 
these prior works focus only on improving the pose estimation accuracy
by using complex and computationally expensive models
whilst largely ignoring the model inference cost issue.
This significantly restricts their scalability and deployability 
in real-world applications particularly 
with very limited computing budgets available.

In the literature, there are a few recent works designed to improve model efficiency.
For example, Bulat and Tzimiropoulos 
built parameter binarised CNN models
to accommodate resource-limited platforms
\cite{bulat2017binarized}.
But this method leads to dramatic performance
drop therefore not satisfied for reliable utilisation.
In most cases, high accuracy rates are required.
Rafi et al. exploited good general purpose practices to improve 
model efficiency without presenting a novel algorithm
\cite{rafi2016efficient}.
Further, this method does not provide quantitative
evaluation on the trade-off between model efficiency and
effectiveness. 

In contrast to these previous methods,
we systematically study the pose estimation efficiency problem
under the condition of preserving the model performance rate
so that the resulted model is more usable and reliable
in real-world application scenarios.

\vspace{0.2cm}
{\bf Knowledge Distillation } 
The objective of knowledge distillation 
is concerned with
information transfer between different neural networks with 
distinct capacities \cite{bucilua2006model,hinton2015distilling,ba2014deep}. 
For instance, 
Hinton et al. successfully employed a well trained large network to help train a small network \cite{hinton2015distilling}.
The rationale is an exploitation of extra supervision from a teacher model, 
represented
in form of class probabilities \cite{hinton2015distilling},
feature representations \cite{ba2014deep,romero2014fitnets},
or an inter-layer flow \cite{yim2017gift}.
This principle has also been recently applied 
to accelerate the model training process of large scale distributed neural networks
\cite{anil2018large},
to transfer knowledge between multiple layers \cite{lan2018person}
or between multiple training states \cite{lan2018self}.
Beyond the conventional two stage training based offline distillation,
one stage online knowledge distillation has been attempted
with added merits of more efficient optimisation
\cite{zhang2018deep,lan2018knowledge}
and more effective learning \cite{lan2018knowledge}.
Besides, knowledge distillation has been exploited 
to distil easy-to-train large networks 
into harder-to-train small networks \cite{romero2014fitnets}.

While these past works above transfer category-level discriminative knowledge, 
our method transfers richer structured information of dense joint confidence maps.
A more similar work is the latest radio signals based pose model
that also adopts the idea of knowledge distillation \cite{zhao2018through}.
However, this method targets at 
using wireless sensors to tackle the occlusion problem,
rather than the model efficiency issue as we confider here.

\section{Fast Human Pose Estimation}
Human pose estimation aims to predict the spatial coordinates of human joints
in a given image.
To train a model in a supervised manner, 
we often have access to a training dataset 
$\{\bm{I}^i,\bm{G}^i\}_{i=1}^{N}$
of $N$ person images each labelled with $K$ joints
defined in the image space as:
\begin{equation}
\bm{G}^i = \{\bm{g}_1^i,..,\bm{g}_K^i\} \in \mathbb{R}^{K \times 2},
\end{equation}
where $H$ and $W$ denotes the image height and width, respectively.
Generally, this is a regression problem at the imagery pixel level.

\vspace{0.2cm}
{\em Objective Loss Function }
For pose model training, we often use
the Mean-Squared Error (MSE) based loss function \cite{tompson2014joint,newell2016stacked}.
To represent the ground-truth joint labels, 
we generate a confidence map ${\bm{m}}_k$ for 
each single joint $k$ ($k \in \{1,\cdots,K\}$)
by centring a Gaussian kernel around the labelled position $\bm{z}_k \!=\! (x_k,y_k)$.

More specifically, a Gaussian  confidence map ${\bm{m}}_k$ for the $k$-th joint label is written as:
\begin{equation}
{\bm{m}}_k(x,y) = \frac{1}{2\pi\sigma^2} \exp\Big({\frac{-[(x-x_k)^2 + (y-y_k)^2]}{2\sigma^2}}\Big)
\end{equation}
where $(x,y)$ specifies a pixel location
and the hyper-parameter $\sigma$ denotes a pre-fixed spatial variance.
The MSE loss function is then obtained as:
\begin{equation}
\mathcal{L}_\text{mse}= 
\frac{1}{K}\sum_{k=1}^{K} \| \bm{m}_k - \hat{\bm{m}}_k \|_2^2
\label{eqn:loss_func}
\end{equation}
where $\hat{\bm{m}}_k$ refers to the predicted confidence map
for the $k$-th joint.
The standard SGD algorithm can then be used to 
optimise a deep CNN pose model by back-propagating 
MSE errors on training data in a mini-batch
incrementally.

Existing pose methods rely heavily on large deep neural networks
for maximising the model performance,
whilst neglecting the inference efficiency.
We address this limitation for higher scalability
by establishing lightweight CNN architectures
and proposing an effective model learning strategy detailed below.

\subsection{Compact Pose Network Architecture}

Human pose CNN models typically consist of
multiple repeated building blocks with the identical structure
\cite{carreira2015human,wei2016convolutional,newell2016stacked,chu2017multi,yang2017pyramid,Nie_2018_CVPR,Peng_2018_CVPR}.
Among these, Hourglass is one of the most common building block units
\cite{newell2016stacked}.
However, we observe that existing designs are not cost-effective,
due to deploying a large number of both channels and blocks
in the entire architecture therefore leading 
to a suboptimal trade-off between 
the representation capability and the computational cost.
For example, \cite{newell2016stacked} suggested a CNN architecture
of 8 Hourglass stages each having 9 Residual blocks with 256 channels within every layer.

We therefore want to minimise the expense of existing CNN architectures 
for enabling faster model inference.
With careful empirical examination,
we surprisingly revealed that a half number of stages (i.e. 4 Hourglass modules)
suffice to achieve over 95\% model generalisation capacity on
the large scale MPII benchmark. 
Moreover, the per-layer channels are also found highly redundant
and reducing a half number (128) only results in
less than 1\% performance drop (Table \ref{tbl:model_reduced}). 
Based on these analysis, we construct a very light CNN architecture
for pose estimation with only one sixth computational cost
of the original design.
See Table \ref{tbl:model_conf} 
and Figure \ref{fig:pipeline} 
for the target CNN architecture specifications.

\begin{table} 
	\centering
	\begin{tabular}{c||c}
		\hline
		Stage & Building Block \\
		\hline \hline
		1, 2, 3, 4 & Hourglass with 128 channels per layer \\
		\hline
	\end{tabular}
	\vskip 0.1cm
	\caption{
		The structure of a small pose CNN model.
	}
	\label{tbl:model_conf}
\end{table}

\vspace{0.2cm}
\textbf{\em Remarks }
Whilst it is attractive to deploy tiny pose networks 
that run cheaply and fast,
it is {\em empirically} non-trivial to train them 
although {\em theoretically} shallow networks 
have the similar representation capacities to 
approximate the target functions as 
learned by deeper counterparts \cite{ba2014deep,seide2011conversational}.
A similar problem has been occurred and investigated in object image classification
through the knowledge distillation strategy, i.e. 
let the target small network mimic the prediction
of a larger teacher model \cite{hinton2015distilling}.
However, it remains unclear how well such a similar method will work
in addressing structured human pose estimation in dense pixel space.
To answer this question, in the following 
we present a pose structure knowledge distillation method.

\subsection{Supervision Enhancement by Pose Distillation}

{\em Model Training Pipeline }
We adopt the generic model training strategy of knowledge distillation:
\begin{enumerate}
	\item We first train a large teacher pose model. In our experiments,
	by default we select the original Hourglass model \cite{newell2016stacked}
	due to its clean design and easy model training.
	Other stronger models can be considered without any restrictions.
	
	\item We then train a target student model with the assistance of
	knowledge learned by the teacher model.
	Knowledge distillation happens in this step.
	The structure of the student model is presented in Table \ref{tbl:model_conf}.
\end{enumerate}

An overview of the whole training procedure is depicted in Figure \ref{fig:pipeline}.
The key to distilling knowledge is 
to design a proper mimicry loss function that is
able to effectively extract and transfer the teacher's knowledge to 
the training of the student model.
The previous distillation function is designed for 
single-label based softmax cross-entropy loss in the context of object categorisation
\cite{ba2014deep,hinton2015distilling} and
unsuitable to transfer the structured pose knowledge
in 2D image space.

To address this aforementioned problem, we design a joint  confidence map dedicated
pose distillation loss function formulated as:
\begin{equation}
\mathcal{L}_\text{pd} = \frac{1}{K} \sum_{k=1}^{K}  
\|\bm{m}^s_{k} - \bm{m}_k^{t}\|_2^2 
\end{equation}
where $\bm{m}^s_{k}$ and $\bm{m}^t_{k}$
specify the confidence maps for the $k$-th joint predicted by
the pre-trained teacher model 
and the in-training student target model, respectively.
We choose the MSE function as the distillation quantity
to measure the divergence between the student and teacher models
in order to maximise the comparability with the pose supervised learning loss
(Eqn \eqref{eqn:loss_func}).

\begin{table*}[h]
	\setlength{\tabcolsep}{0.1cm}
	\begin{center}
				\resizebox{\textwidth}{!}{
		\begin{tabular}{ c || c |c | c | c | c | c | c || c | c|| c | c }
			\hline
			Method &Head & Sho. & Elbo. & Wri. & Hip & Knee  & Ank. & Mean & AUC & \# Param  &  Deployment Cost \\ 
			\hline \hline
			Rafi et al., BMVC'16\cite{rafi2016efficient}
			& 97.2  & 93.9  & 86.4  & 81.3  & 86.8  & 80.6 & 73.4 & 86.3 & 57.3 
			& 56M 
			& 28G \\  \hline
			Belagiannis\&Zisserman, FG'17\cite{belagiannis2016recurrent}
			& 97.7  & 95.0  & 88.2  & 83.0  & 87.9  & 82.6 & 78.4 & 88.1 & 58.8 
			& 17M 
			& 95G \\ \hline
			Insafutdinov et al., ECCV'16\cite{insafutdinov2016deepercut}
			& 96.8  & 95.2  & 89.3  & 84.4  & 88.4  & 83.4 & 78.0 & 88.5 & 60.8 
			& 66M 
			& 286G \\ \hline
			Wei et al., CVPR'16\cite{wei2016convolutional}
			& 97.8  & 95.0  & 88.7  & 84.0  & 88.4  & 82.8 & 79.4 & 88.5 & 61.4 
			& 31M 
			& 351G \\ \hline
			Bulat\&Tzimiropoulos, ECCV'16\cite{bulat2016human}
			& 97.9  & 95.1  & 89.9  & 85.3  & 89.4  & 85.7 & 81.7 & 89.7 & 59.6 
			& 76M 
			& 67G \\ \hline
			Newell et al., ECCV'16\cite{newell2016stacked}
			& 98.2  & 96.3  & 91.2  & 87.1  & 90.1  & 87.4 & 83.6 & 90.9 & 62.9 
			& 26M 
			& 55G \\ \hline
			Ning et al., TMM'17\cite{ning2017tmm}
			& 98.1  & 96.3  & 92.2  & 87.8  & 90.6  & 87.6 & 82.7 & 91.2 & 63.6 
			& 74M 
			& 124G \\ \hline
			Chu et al., CVPR'17\cite{chu2017multi}
			& 98.5  & 96.3  & 91.9  & 88.1  & 90.6  & 88.0 & 85.0 & 91.5 & 63.8 
			& 58M 
			& 128G \\ \hline
			Peng et al., CVPR'18{\cite{Peng_2018_CVPR}}
			& 98.1  & 96.6  & 92.5  & 88.4   & 90.7 & 87.7 & 83.5 & 91.5 & -
			& 26M 
			& 55G \\ \hline
			Yang et al., ICCV'17\cite{yang2017pyramid}
			& 98.5  & 96.7  &  92.5 & 88.7  & 91.1 &  88.6 & 86.0 &  92.0  & 64.2 
			& 28M 
			& 46G  \\ \hline
			Nie et al., CVPR'18{\cite{Nie_2018_CVPR}}
			& \bf98.6 & \bf96.9 & \bf93.0 & \bf89.1 & \bf91.7 & \bf89.0 & \bf86.2 & \bf92.4 & \bf65.9 
			& 26M & 
			63G \\  \hline 
			Sekii, ECCV18{\cite{Sekii2018ECCV}} 
			& -  & -  & -  & -  & -  & - & - & 88.1 & - 
			& 16M & \bf 6G \\
			\hline \hline
			\bf  
			FPD 
			& 98.3  & 96.4  & 91.5  & 87.4  & 90.9  & 87.1 & 83.7 & 91.1 & 63.5 
			& \bf 3M 
			& 9G \\ 
			
			\hline
		\end{tabular}
		}
	\end{center}
	\vskip 0.1cm
	\caption{
		PCKh@$0.5$ and AUC (\%) rates on the MPII {\em test} dataset.
		M/G: $10^6$/$10^9$. 
	} 
	\label{tbl:performance_mpii}
\end{table*}

\vspace{0.2cm}
{\bf Overall Loss Function }
We formulate the overall FPD loss function for pose structure knowledge distillation during training as:
\begin{equation}
\mathcal{L}_\text{fpd} = \alpha \mathcal{L}_\text{pd} 
+ (1-\alpha) \mathcal{L}_\text{mse} 
\label{eqn:FPD_loss} 
\end{equation}
where $\alpha$ is the balancing weight between 
the two loss terms, estimated by cross-validation.
As such, the target network learns both to predict the labelled ground-truth annotations of training samples by $\mathcal{L}_\text{mse}$ 
and to match the prediction structure of the stronger teacher model by $\mathcal{L}_\text{pd}$.

\vspace{0.2cm}
\textbf{\em Further Remarks }
Why does the proposed pose distillation loss function 
probably help to train a more generalisable target model,
as compared to training {\em only} on the labelled data?
A number of reason may explain this in the context of pose estimation.
\begin{enumerate}
	\item The body joint labels are likely to be erroneous due to the high 
	difficulty of locating the true positions in the manual annotation process.
	In such cases, the teacher model may be able to mitigate some errors
	through statistical learning and reasoning therefore reducing the misleading effect 
	of wrongly labelled training samples
	(Figure \ref{fig:vis_example} Row (A)).
	
	\item Given difficult training cases say with confusing/cluttered background
	and random occlusion situations,
	the teacher prediction may provide softened learning tasks
	by explained away these hard samples
	with model inference (Figure \ref{fig:vis_example} Row (B)).

	\item The teacher model may provide more complete joint labels
	than the original annotation therefore not only
	providing additional more accurate supervision
	but also mitigating the misleading of missing joint labels 
	(Figure \ref{fig:vis_example} Row (C)).
	
	\item Learning to match the ground-truth  confidence map can be harder 
	in comparison to aligning the teacher's prediction.
	This is because the teacher model has spread some reasoning uncertainty
	for each training sample either hard or easy to process.
	
	\item On the other hand, the teacher's  confidence map encodes the abstract knowledge 
	learned from the entire training dataset in advance, 
	which may be beneficial to be considered
	in learning every individual training sample during knowledge distillation.
\end{enumerate}

In summary, the proposed model is capable of handling wrong pose joint annotations, e.g. when the pre-trained teacher predicts more accurate joints than manual wrong and missing labels. Due to a joint use of the ground-truth labels and the teacher model's prediction, our model is tolerant to either error but not co-occurring ones. 
This alleviates the harm of label errors in the training data, in contrast to existing methods that often blindly trust all given labels.

\subsection{Model Training and Deployment}
The proposed FPD model training method consists of two stages:
(i) We train a teacher pose model by the conventional MSE loss (Eqn  \eqref{eqn:loss_func}),
and (ii) train a target student model by the proposed
loss (Eqn \eqref{eqn:FPD_loss}),
with the knowledge distillation from the teacher model to
the target model being conducted in each mini-batch and throughout the entire training process.
At test time, we only use the small target model
for efficient and cost-effective deployment 
whilst throwing away the heavy teacher network.
The target model already extracts the teacher's knowledge.

\section{Experiments}
\subsection{Experiment Setup}

{\bf Datasets }
We utilised two human pose benchmark datasets,
MPII \cite{andriluka14cvpr} and
Leeds Sports Pose (LSP) \cite{johnson2010clustered}.
The MPII dataset is collected from YouTube videos 
with a wide range of human activities and events.
It has 25K scene images and 40K annotated persons (29K for
training and 11K for test). 
Each person has 16 labelled body joints.
We adopted the standard train/valid/test data split \cite{tompson2015efficient}.
Following \cite{tompson2014joint}, we randomly sampled 3K
samples from the training set for model validation. 

The LSP benchmark contains natural person images from many different sports
scenes. Its extended version provides 11K training samples and
1K test samples. Each person in LSP has 14 labelled joints.

\begin{table*} 
	\setlength{\tabcolsep}{0.1cm}
	\begin{center}
		\resizebox{\textwidth}{!}{%
			\begin{tabular}{c || c |c | c | c | c | c | c || c | c || c | c }
				\hline
				Method & Head & Sho. & Elbo. & Wri. & Hip & Knee  & Ank. & Mean & AUC &\# Param & Deployment Cost  \\
				\hline \hline
				Tompson et al., NIPS'14\cite{tompson2014joint}
				& 90.6  & 79.2  & 67.9  & 63.4  & 69.5  & 71.0 & 64.2 & 72.3 & 47.3 
				& - & - \\ \hline
				Fan et al., CVPR'15\cite{fan2015combining}
				& 92.4  & 75.2  & 65.3  & 64.0  & 75.7  & 68.3 & 70.4 & 73.0 & 43.2 
				& - & - \\ \hline
				Carreira et al., CVPR'16\cite{carreira2015human}
				& 90.5  & 81.8  & 65.8  & 59.8  & 81.6  & 70.6 & 62.0 & 73.1 & 41.5 
				& - & - \\ \hline
				Chen\&Yuille, NIPS'14\cite{chen2014articulated}
				& 91.8  & 78.2  & 71.8  & 65.5  & 73.3  & 70.2 & 63.4 & 73.4 & 40.1 
				& - & - \\ \hline
				Yang et al., CVPR'16\cite{yang2016end}
				& 90.6  & 78.1  & 73.8  & 68.8  & 74.8  & 69.9 & 58.9 & 73.6 & 39.3 
				& - & - \\ \hline
				Rafi et al., BMVC'16\cite{rafi2016efficient}
				& 95.8  & 86.2  & 79.3  & 75.0  & 86.6  & 83.8 & 79.8 & 83.8 & 56.9 
				& 56M  
				& 28G \\ \hline
				Yu et al., ECCV'16\cite{yu2016deep}
				& 87.2  & 88.2  & 82.4  & 76.3  & 91.4  & 85.8 & 78.7 & 84.3 & 55.2 
				& - & - \\ \hline 
				\hline
				Peng et al., CVPR'18\cite{Peng_2018_CVPR}
				& \bf 98.6 & \bf 95.3 & \bf 92.8 & \bf 90.0 & \bf 94.8 & \bf 95.3 & \bf 94.5 & \bf 94.5 & -
				&   26M  
				&  55G \\ \hline \hline
				\bf FPD
				& 97.3  & 92.3  & 86.8  & 84.2  & 91.9  & 92.2 & 90.9 & 90.8 & \bf 64.3
				&  \bf 3M  
				& \bf 9G     
				\\ \hline
			\end{tabular}
		}
	\end{center}
	\vskip 0.1cm
	\caption{
		PCK@$0.2$ and AUC (\%) rates on the LSP {\em test} dataset. 
		M/G: $10^6$/$10^9$.
	}
	\label{tbl:performance_lsp}
\end{table*}

\vspace{0.2cm}
{\bf Performance Metrics }
We used the standard
Percentage of Correct Keypoints (PCK) measurement
that quantifies the fraction of correct predictions 
within an error threshold $\tau$
\cite{yang2013articulated}.
Specifically, the quantity $\tau$ is normalised against the size of either torso 
($\tau\!=\!0.2$ for LSP, i.e. PCK@$0.2$) or head
($\tau\!=\!0.5$ for MPII, i.e. PCKh@$0.5$).
We measured each individual joint respectively
and took their average as an overall metric.
Using different $\tau$ values, we yielded a PCK curve.
Therefore, the Area Under Curve (AUC) can be obtained as a holistic measurement
across different decision thresholds.
To measure the model efficiency both in training and test,
we used the FLOPs.

\vspace{0.2cm}
{\bf Training Details }
We implemented all the following experiments
in Torch.
We cropped all the training and test images according to the provided positions and scales, and resized them to 256$\times$256 in pixels.
As typical,
random scaling (0.75-1.25), rotating ($\pm$30 degrees) and 
horizontal flipping were performed to augment the training data.
We adopted the RMSProp optimisation algorithm.
We set the learning rate to $2.5\!\times \! 10^{-4}$,
the mini-batch size to 4,
and the epoch number to 130 and 70 for MPII and LSP benchmarks, respectively.
For the network architecture,
we used the original Hourglass as the teacher model
and the customised Hourglass with
less depth and width (Table \ref{tbl:model_conf}) as the target model.

\begin{figure*}
	\centering
	\includegraphics[width=1\linewidth]{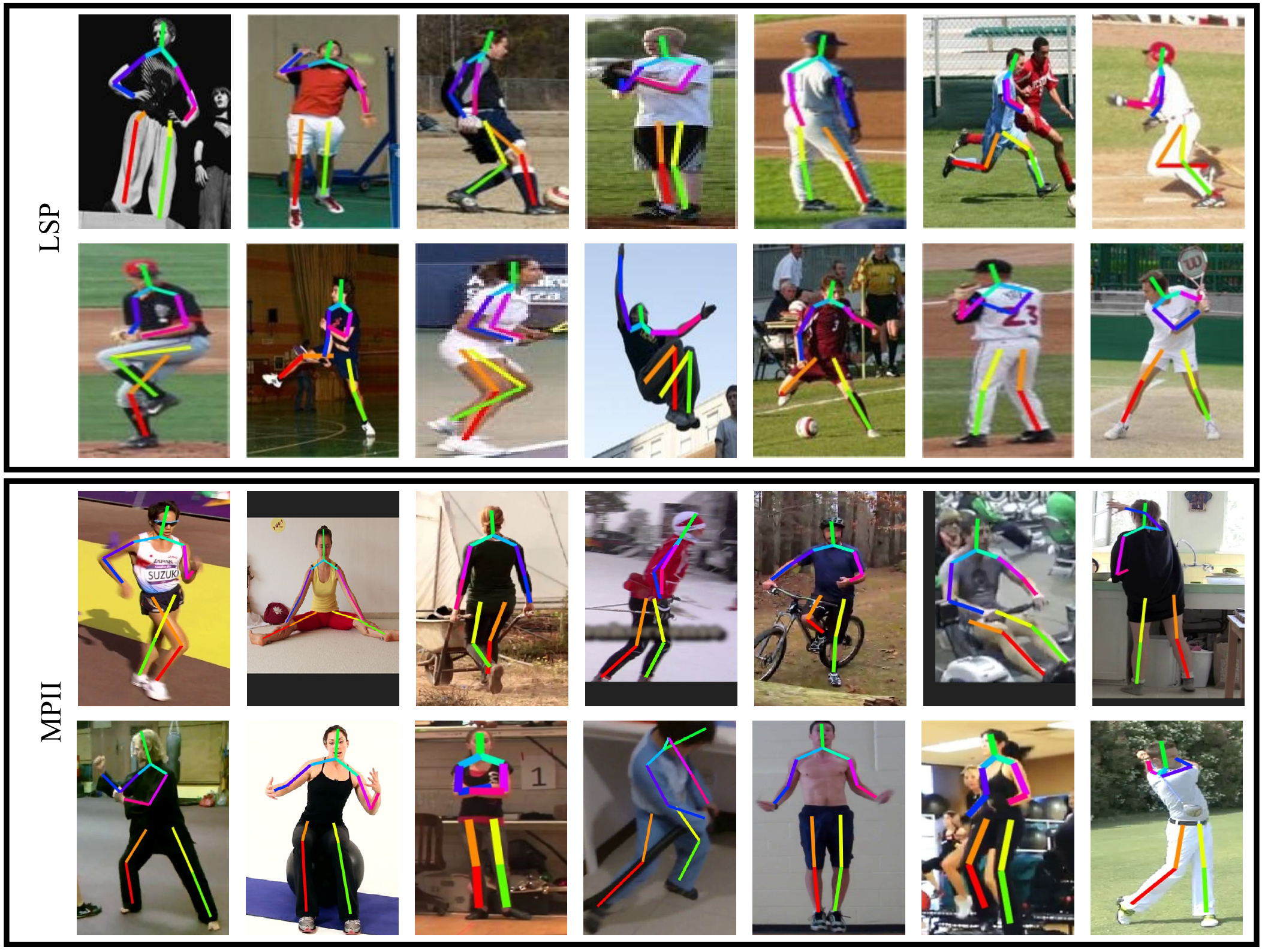}
	\vskip 0.1cm
	\caption{
		Example of human pose estimation on LSP and MPII.
	}
	\label{fig:vis_lspmpii_samples}
\end{figure*}

\begin{figure*} 
	\centering
	\includegraphics[width=1\linewidth]{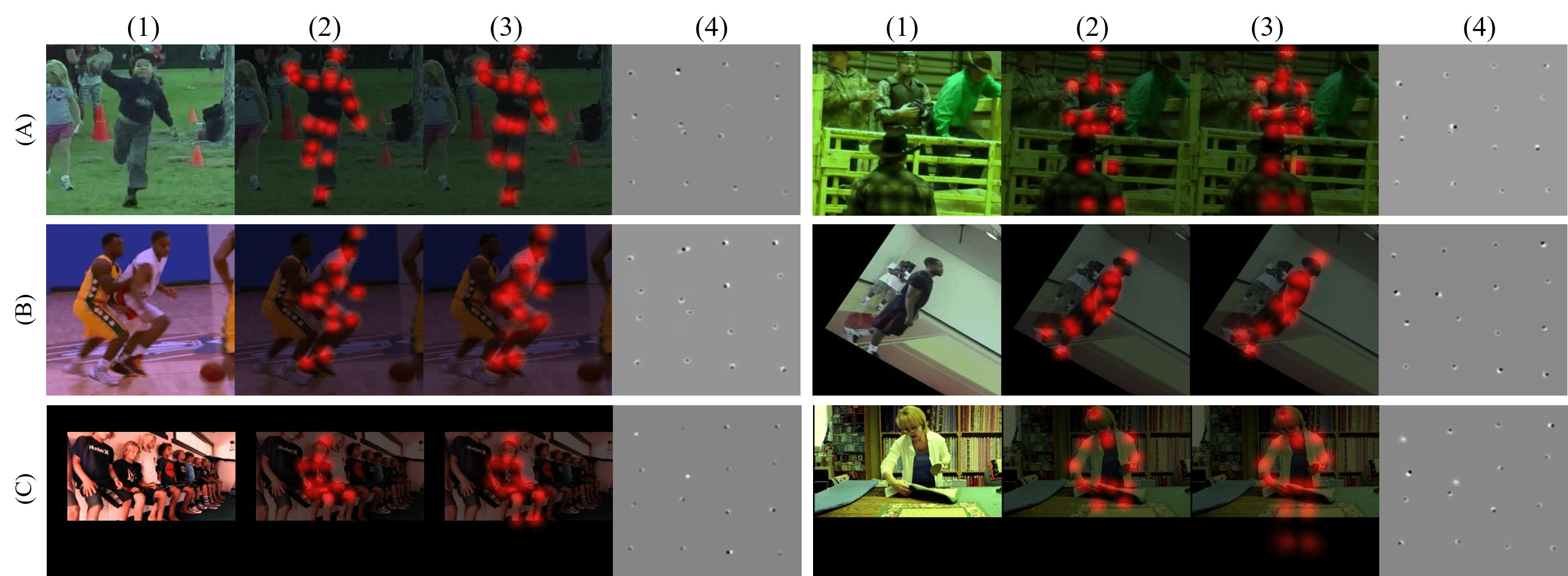}
	\vskip 0.1cm
	\caption{
		Pose estimation examples on MPII by the proposed FPD model.
		Column {\bf (1)}: The input images.
		Column {\bf (2)}: Ground-truth joint confidence maps.
		Column {\bf (3)}: Joint confidence maps predicted by the teacher model.
		Column {\bf (4)}: The difference between ground-truth and teacher's  confidence map.
		Each row represents a type of pose knowledge transfer.
		Row {\bf (A)}: Error labelling of the right leg ankle in the ``ground-truth'' annotations,
		which is corrected by the teacher model.
		Row {\bf (B)}: A softened teacher confidence map with larger uncertainty 
		than the ground-truth due to the highly complex 
		human posture. 
		Row {\bf (C)}: Missing joint labels are discovered by the teacher model. 
	}
	\label{fig:vis_example}
\end{figure*}

\begin{table*}
	\setlength{\tabcolsep}{0.35cm}
	\begin{center}
		\begin{tabular}{ c ||c| c | c | c | c | c | c || c |c }
			\hline
			FPD &  Head & Sho. & Elbo. & Wri. & Hip & Knee  & Ank. & Mean & AUC  \\
			\hline \hline
			\xmark 
			& 97.4 & 96.0 & 90.2 & 85.8 & 88.2 & 84.3 & 80.6 & 89.4 & 61.4
			\\ \hline
			\cmark
			& \bf 97.5 &\bf  96.3 &\bf  91.4 &\bf  87.3 &\bf  89.4 &\bf  85.6 &\bf  82.0 &\bf  90.4 &\bf 62.4 
			\\
			\hline
		\end{tabular}
	\end{center}
	\vskip 0.1cm
	\caption{
		Generalisation evaluation of the proposed FPD approach. 
		{Metric}: Mean PCKh@$0.5$ and AUC.
	}
	\label{tbl:generalisation_test}
\end{table*}

\begin{table*} 
	\setlength{\tabcolsep}{0.7cm}
	\centering 
	\begin{tabular}{ c | c || c | c || c | c }
		\hline
		\# Stage & \# Channel & Mean & AUC & \# Param & Deployment Cost \\
		\hline \hline
		8 & 256 & 91.9 & 63.7 & 26M 
		& 55G \\ 
		\hline
		4 & 256 & 91.4 & 63.9  & 13M 
		& 30G \\ 
		\hline
		2 & 256 & 90.5 & 63.0 & 7M 
		& 17G \\ 
		\hline
		1 & 256 & 86.4 & 58.3 & 3M 
		& 10G \\
		\hline 
		\hline
		
		4 & 256 & 91.4 & 63.9  & 13M 
		& 30G \\ 
		\hline
		4 & 128 & 90.1 & 62.4 & 3M 
		& 9G \\ 
		\hline 
		4 & 64 & 87.9 & 59.5 & 0.95M 
		& 4.5G        \\ 
		\hline
		4 & 32  & 83.4 & 54.9 & 0.34M 
		& 3.1G        \\ 
		\hline
	\end{tabular}
	\vskip 0.1cm
	\caption{
		Cost-effectiveness analysis of the Hourglass model.
		{Metric}: PCKh@$0.5$ and AUC.
		M/G: $10^6$/$10^9$.
	}
	\label{tbl:model_reduced}
\end{table*}

\begin{table} [h]
	\setlength{\tabcolsep}{0.5cm}
	\centering
	\begin{tabular}{ l || c | c }
		\hline
		Pose Distillation &  Mean  &  AUC  \\
		\hline \hline
		\xmark 
		& 90.1 & 62.4
		\\
		\hline
		\cmark 
		&\bf 90.9 &\bf 63.3
		\\ 
		\hline
		%
	\end{tabular}
	\vskip 0.1cm
	\caption{
		Effect of the proposed pose knowledge distillation.
		{Metric}: Mean PCKh@$0.5$ and AUC (\%). 
	}
	\label{tbl:effect_KD}
\end{table}

\begin{table*}
	\setlength{\tabcolsep}{0.35cm}
	\begin{center}
		\begin{tabular}{ c ||c| c | c | c | c | c | c || c |c }
			\hline
			Loss Function &  Head & Sho. & Elbo. & Wri. & Hip & Knee  & Ank. & Mean & AUC  \\
			\hline \hline
			MSE&  \bf97.7  & \bf 96.4  & \bf 91.8  & 87.6  & \bf89.7  &\bf 86.6 &\bf 83.9 &  \bf 90.9  & \bf 63.3       \\
			\hline
			
			
			Cross-Entropy   &  97.6  & 96.2  & 91.5  & 87.6  & 89.0  & 86.5 & 83.6 & 90.7 & 63.0 \\
			\hline
		\end{tabular}
	\end{center}
	\vskip 0.1cm
	\caption{
		Pose knowledge distillation by different types of loss function. 
		{Metric}: Mean PCKh@$0.5$ and AUC. 
	}
	\label{tbl:kdlosstype}
\end{table*}

\subsection{Comparisons to State-Of-The-Art Methods}

We evaluated the proposed FPD method by extensively
comparing against recent human pose estimation deep methods
on MPII and LSP.

\vspace{0.2cm}
{\bf Results on MPII }
Table \ref{tbl:performance_mpii} compares the PCKh@$0.5$
accuracy results of state-of-the-art methods and the proposed FPD on 
the {\em test} dataset of MPII.
It is clearly observed that the proposed FPD model
is significantly efficient and compact therefore 
achieving a much cheaper deployment cost.
Importantly, this advantage is obtained {\em without} clearly 
compromising the model generalisation capability,
e.g. achieving as high as 91.1\%.

Specifically, compared with the best performer \cite{Nie_2018_CVPR},
the FPD model only requires $14.3\%$ (9/63) computational cost 
but gaining 96.4$\%$ (63.5/65.9) performance
in mean PCKh accuracy.
This leads to a 6.7\%$\times$ (96.4/14.3) cost-effective advantage.
When compared to the most efficient alternative competitor \cite{rafi2016efficient},
our model is 2.9$\times$ (26/9) more efficient
whilst simultaneously achieving a mean PCKh gain of 4.8$\%$ 
(91.1-86.3).
These evidences clearly suggest the cost-effectiveness advantages of our method
over other alternative approaches.

In pose estimation, an improvement of 0.8\% indicates a significant gain particularly on the challenging MPII with varying poses against cluttered background. This boost is bigger than other state-of-the-art gains, 
e.g. +0.3\% in 91.2\% \cite{ning2017tmm} vs 90.9\% \cite{newell2016stacked}; 
further +0.3\% in 91.5\% \cite{peng2018jointly}. 
More specifically, given all 163,814 test joints, each 0.1\% gain means correcting 163 joints.

\vspace{0.2cm}
{\bf Results on LSP }
Table \ref{tbl:performance_lsp} compares the PCK@$0.2$ rates of 
our FPD model and existing methods 
with top reported performances on the LSP test data.
Compared to MPII,
this benchmark has been less evaluated by deep learning models,
partly due to a smaller size of training data.
Overall, we observed the similar comparisons.
For example, our FPD runs more efficiently
than the most competitive alternative \cite{rafi2016efficient} and 
consumes much less training energy,
in addition to achieving the best pose prediction
accuracy rate among all compared methods.

\vspace{0.2cm}
{\bf Qualitative Examination }
To provide visual test, Figure \ref{fig:vis_lspmpii_samples} shows qualitative pose estimation evaluations on LSP and MPII. 
It is observed that such a small FPD model 
can still achieve reliable and robust pose estimation in 
arbitrary in-the-wild images with various background clutters, different human poses
and viewing conditions.


\begin{table} 
	\setlength{\tabcolsep}{0.23cm}
	\centering
	\begin{tabular}{ c ||c|c|c|c|c|c}
		\hline
		$\alpha$ &\em 0 &\em 0.05 &\em  0.1 &\em 0.5 &\em 0.95 &\em 0.99 \\
		\hline \hline
		Mean & 90.1 & 90.8 & 90.8 &\bf 90.9 & 90.7 & 90.7 \\
		\hline
		AUC & 62.4 & 63.2 & 63.2 &\bf 63.3 & 63.0 & 63.0 \\
		\hline
	\end{tabular}
	\vskip 0.1cm
	\caption{
		Performance analysis of the learning importance parameter of pose distillation.
		{Metric}: Mean PCKh@$0.5$ and AUC (\%).
	}
	\label{tbl:alpha}
\end{table}

\subsection{Ablation Study}

We carried out detailed component analysis and discussion
on the {\em validation} set of MPII.

\vspace{0.2cm}
{\bf FPD generalisation evaluation }
Besides using the state-of-the-art Hourglass as the backbone network,
we also tested the more recent model \cite{yang2017pyramid} when 
integrated into the proposed FPD framework.
In particular, we adopted the original network as the teacher model 
and constructed a lightweight variant as the student (target) model.
The lightweight model was constructed similarly as in Table \ref{tbl:model_conf}
because it is based on the Hourglass design too: 
reducing the number of stages to 4 and 
the number of channels in each module to 128.
The results in Table \ref{tbl:generalisation_test}
show that our FPD approach achieves 1.0\% mean PCKh@$0.5$ gain,
similar to the Hourglass case.
This suggests the good generalisation capability of the proposed approach
in yielding cost-effective pose estimation deep models.

\vspace{0.2cm}
{\bf Cost-effectiveness analysis of Hourglass }
We extensively examined the architecture design 
of the state-of-the-art Hourglass neural network model \cite{newell2016stacked}
in terms of cost-effectiveness. 
To this end, we tested two dimensions in design:
depth (the layer number) and width (the channel number).
Interestingly, we revealed in Table \ref{tbl:model_reduced} that removing half stages (layers)
and half channels only leads to quite limited performance degradation.
This indicates that the original Hourglass design is highly
redundant with poor cost-effectiveness.
However, this is largely ignored in previous works
due to their typical focus on pursuing the model accuracy performance
{\em alone} whilst overlooking the important model efficiency problem.
This series of CNN architecture examinations 
helps us to properly formulate a lightweight pose CNN architecture
with only 16\% (9/55) computational cost but obtaining
98\% (90.1/91.9) model performance as compared to the state-of-the-art design,
laying a good foundation towards building compact yet strong 
human pose deep models.

\vspace{0.2cm}
{\bf Effect of pose knowledge distillation }
We tested the effect of using our pose knowledge distillation
on the lightweight Hourglass network. 
In contrast to all other methods, the model \cite{peng2018jointly} additionally benefits from an auxiliary dataset MPII in model training.
Table \ref{tbl:effect_KD} shows that teacher knowledge transfer
brings in 0.8\% (90.9-90.1) mean PCKh accuracy boost.
This suggests that the generic principle of knowledge distillation 
is also effective in the structured pose estimation context,
beyond object categorisation.

To further validate how on earth this happens, we visualise three pose structure
transfer examples in Figure \ref{fig:vis_example}. 
It is shown that the proposed mimicry loss 
against the teacher prediction is likely to pose extra information
in cases of error labelling, hard training images, and missing annotation.

\vspace{0.2cm}
{\bf Pose distillation loss function }
We finally evaluated the effect of loss function choice for pose knowledge distillation.
To that end, we further tested a Cross-Entropy measurement based
loss. Specifically, we first normalise the entire  confidence map so that
the sum of all pixel confidence scores is equal to 1, i.e. L1 normalisation.
We then measure the divergence between the predicted and ground-truth
confidence maps using the Cross-Entropy criterion.
The results in Table \ref{tbl:kdlosstype}
show that the MSE is a better choice in comparison to Cross-Entropy.
The plausible reason is that MSE is also the formulation of the conventional 
supervision loss (Eqn \eqref{eqn:loss_func}) therefore more compatible.

\vspace{0.2cm}
{\bf Parameter analysis of loss balance }
We evaluated the balance importance between the conventional MSE loss
and the proposed pose knowledge distillation loss,
as controlled by $\alpha$ in Eqn \eqref{eqn:FPD_loss}.
Table \ref{tbl:alpha} shows that equal importance (when $\alpha\!=\!0.5$)
is the optimal setting. This suggests that the two loss terms are 
similarly significant with the same numerical scale.
On the other hand, we found that this parameter setting is not sensitive
with a wide range of satisfactory values.
This indicates that the teacher signal is not far away from 
the ground-truth labels (see Figure \ref{fig:vis_example} Column (4)),
possibly providing an alternative supervision
as a replacement of the original joint  confidence map labels.

\section{Conclusion}
In this work, we present a novel 
{\em Fast Pose Distillation} (FPD) learning strategy.
In contrast to most existing human pose estimation methods,
the FPD aims to address the under-studied and 
practically significant model cost-effectiveness quality
in order to scale the human pose estimation models
to large deployments in reality.
This is made possible by developing a lightweight
human pose CNN architecture and 
designing an effective pose structure knowledge distillation
method from a large teacher model
to a lightweight student model.
Compared with existing model compression techniques
such as network parameter binarisation,
the proposed method achieves highly efficient 
human pose models {\em without} accuracy performance compromise.
We have carried out extensive comparative evaluations
on two human pose benchmarking datasets.
The results suggests the superiority of our FPD approach 
in comparison to a wide spectrum of state-of-the-art 
alternative methods.
Moreover, we have also conducted a sequence of ablation study
on model components to provide detailed analysis and insight about 
the gains in model cost-effectiveness.

\section{Acknowledgement}
This work was supported in part by the National Natural Science Foundation of China (61773093), National Key R\&D Program of China (2018YFC0831800), Important Science and Technology Innovation Projects in Chengdu (2018-YF08-00039-GX) and Research Programs of Sichuan Science and Technology Department (2016JY0088, 17ZDYF3184). Mao Ye is the major corresponding author.


{\small
\bibliographystyle{ieee_fullname}
\bibliography{cvpr2019}
}

\end{document}